\documentclass{article}

\PassOptionsToPackage{numbers, square, compress}{natbib}
 \usepackage[preprint]{neurips_2026}

\usepackage[utf8]{inputenc} 
\usepackage[T1]{fontenc}    
\usepackage{hyperref}       
\usepackage{url}            
\usepackage{booktabs}       
\usepackage{amsfonts}       
\usepackage{nicefrac}       
\usepackage{microtype}      
\usepackage{xcolor}         
\usepackage{multirow}
\usepackage{graphicx}
\usepackage{wrapfig}        
\usepackage{amsmath}
\usepackage{cleveref}
\usepackage{amsmath}
\usepackage{booktabs, multirow, graphicx}
\usepackage{bbm}

\title{Drift-Resistant Navigation World Model with Anchored Epipolar Guidance}

\author{%
{\bfseries Po-Chien Luan \hspace{2.2em} Zimin Xia$^\dagger$ \hspace{2.2em} Wuyang Li \hspace{2.2em} Yang Gao \hspace{2.2em} Alexandre Alahi}$^\dagger$\\[0.1cm]
{\normalfont VITA@EPFL} \\
\texttt{firstname.lastname@epfl.ch} \hspace{1.2em} $^\dagger$ Corresponding Author\\
\\
{\normalfont\small Project Page:
\hypersetup{urlcolor=magenta}
\href{https://vita-epfl.github.io/DR-NWM.io/}{\texttt{\textcolor{magenta}{https://vita-epfl.github.io/DR-NWM.io/}}}}%
}

\begin{document}

\maketitle


\begin{abstract}
We propose Drift-Resistant Navigation World Model, a generative model that mitigates both \emph{perceptual drift} and \emph{geometric drift} in conventional rollout-based navigation world models. Existing methods recursively feed generated content into subsequent steps, causing noise accumulation and degraded predictions, \emph{i.e.}, \emph{perceptual drift}.
Meanwhile, their predictions often deviate from the agent’s motion, resulting in \emph{geometry drift}.
We address both types of drift by redesigning world-model prediction as an \emph{anchor-guided rollout}. Instead of rolling out every frame sequentially, we first predict sparse future anchors that serve as stable long-range targets, and then generate intermediate frames within each chunk conditioned on both past context and future anchors. Importantly, these sparse anchors also provide geometric constraints, supported by bidirectional epipolar geometry, to localize where corresponding content should appear in the intermediate frames.
Experiments on four benchmarks demonstrate consistent improvements over strong baselines in long-horizon visual quality, geometric consistency, and multi-view coherence. These gains further translate into improved downstream planning performance under the same planners, highlighting the importance of drift-resistant, geometry-aware prediction for reliable navigation world models.

\end{abstract}
\section{Introduction}

World models \cite{ha2018recurrent} learn the dynamics of the environment, enabling an agent to anticipate future observations before acting \cite{lecun2022path,hafner2023mastering,agarwal2025cosmos}. Navigation is a fitting setting for such prediction \cite{shah2023vint,bar2025navigation}. Because each action changes the robot’s viewpoint, navigation depends on predicting what it will observe after moving. Therefore, a world model provides a natural basis for navigation by forecasting future observations under candidate actions, which can then be used to compare possible trajectories.

\begin{figure}[tb]
\centering

  \includegraphics[width=1.0\linewidth]{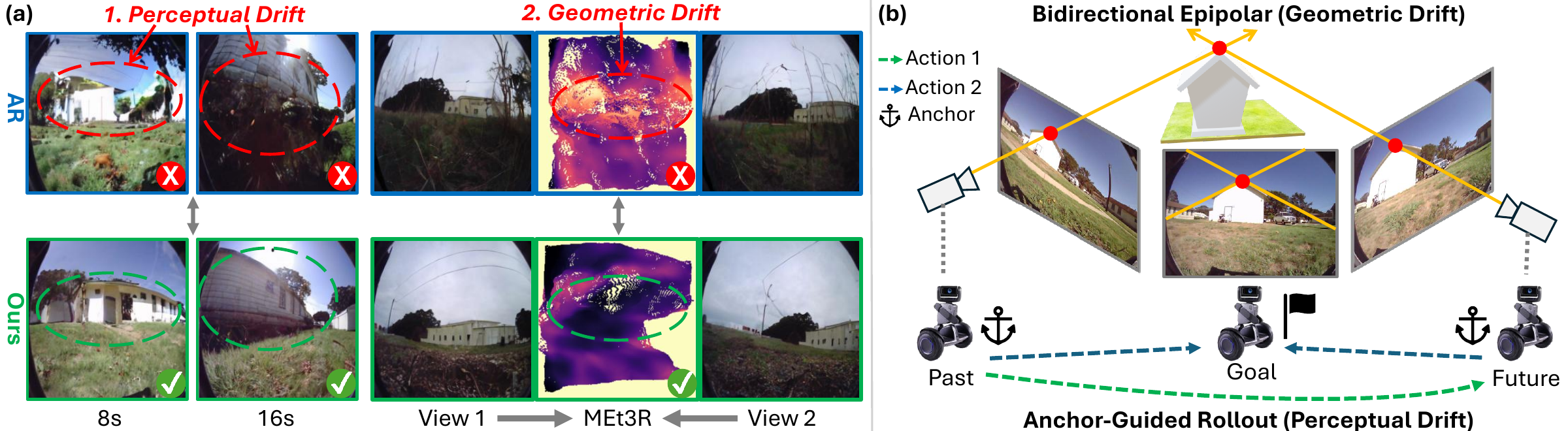}

\caption{
(a) Our method improves perceptual and geometric drifts (left: visual quality in long horizons; right:  MEt3R \cite{asim2025met3r} (darker is better)). We use NWM \cite{bar2025navigation} as an example of an autoregressive (AR) process. Red circles denote the drifts.
(b) Anchor-guided rollout improves perceptual drift. Through two anchors, we can form bidirectional epipolar constraints to improve geometric drift.
}
\label{fig:pull} 
\end{figure}

Despite their promise, current world models for navigation suffer from two drifts (\cref{fig:pull}a). The first is \textbf{\emph{perceptual drift}} \cite{yan2026mwm, ren2025gen3c, li2025stable, li2026everanimate}. Standard practice~\cite{bar2025navigation} involves forecasting future autoregressively, where previous predictions are recursively fed back as inputs for subsequent steps. Consequently, minor per-step errors propagate and compound over time, leading to perceptual drift over extended horizons~\cite{bengio2015scheduled}. While some methods attempt to mitigate this through inference-like rollout training \cite{yan2026mwm}, exposing the model to its own errors in this manner introduces prohibitive computational overhead.

The second is \textbf{\emph{geometric drift}} \cite{yan2026mwm, wang2024motionctrl, bagchi2026walk}. There are two main reasons for this drift. First, predicted futures may fail to stay aligned with the changes of the agent’s actions, even when they remain perceptually plausible. Second, perceptual drift further corrupts geometric consistency, since degraded predictions provide unreliable visual cues for subsequent viewpoint changes. To enforce geometric fidelity, prior approaches typically rely on costly 3D supervision, such as depth maps or point clouds \cite{kupyn2025epipolar, wu2025geometry, dai2025fantasyworld, zhang2026rae}. Such supervision is difficult to acquire at scale and often limits generalizability. Both failures are critical bottlenecks for navigation: effective planning requires future simulations that remain stable over long horizons and faithful to the geometry induced by the agent’s motion.

To address these challenges, we adopt a strategy mirrored in both classical robotics \cite{latombe2012robot} and human behavior \cite{miller2017plans}: first planning a distant goal and then determining the steps to reach it. Following this intuition, we decompose future prediction into a two-level hierarchy. Our model first establishes a series of sparse temporal \emph{anchors} that act as long-term goals for the trajectory. Once these stable anchors are fixed, the model tracks the path between them by generating intermediate frames for each chunk. This effectively partitions the full rollout into shorter, manageable chunks, preventing recursive error accumulation and fundamentally mitigating perceptual drift (\cref{fig:pull}a left).

Furthermore, the anchors provide not only temporal guidance but also a geometric benefit. Once both the past and future anchors are available, they define two views of the same rollout, while the action sequence between them provides a motion prior for the intermediate frames. Based on these two views, we use bidirectional epipolar geometry to localize where corresponding content should appear in the intermediate frames (proof in app.~\ref{app:proof_epi}), as illustrated in \cref{fig:pull}(b). This gives the model stronger spatial grounding without requiring explicit 3D supervision (\cref{fig:pull}a right).

We propose \textbf{Drift-Resistant Navigation World Model (DR-NWM)}, a generative model that combines \emph{anchor-guided rollout} and \emph{bidirectional epipolar geometry} with a  \emph{Anchor-conditioned Diffusion Transformer} (AC-DiT). DR-NWM first predicts sparse anchors at fixed intervals, then generates intermediate frames within each chunk by jointly conditioning on past context and the future anchor. By leveraging bidirectional actions and views, we inject geometry awareness into AC-DiT through bidirectional epipolar masks. As a result, the generated sequences not only remain stable over time but also stay faithful to the geometry induced by the agent’s motion.

This paper makes two main contributions. 1) We propose DR-NWM. It jointly mitigates perceptual drift and geometric drift, leading to consistent improvements in long-horizon future prediction and multi-view consistency. These improvements further translate into stronger downstream planning performance. 2) At the core of DR-NWM is {anchor-guided rollout}, our ingenious design that redefines the role of future anchors: they serve not only as stable long-horizon targets for mitigating recursive drift, but also as key geometric anchors for bidirectional epipolar grounding, enabling geometry-aware prediction without explicit 3D supervision.

\section{Related Work}
Recent work has explored world models as predictive simulators for navigation, improving the paradigm along several directions. NWM establishes controllable future-view generation for navigation with a Conditional Diffusion Transformer backbone, following a conventional autoregressive rollout pipeline that predicts future observations sequentially from past observations and actions \cite{bar2025navigation}. While effective, this design remains vulnerable to long-horizon error accumulation. MWM improves rollout consistency and deployment efficiency \cite{yan2026mwm}, while RAE-NWM studies navigation modeling in dense visual representation space to improve long-horizon behavior and action accuracy \cite{yan2026mwm}. UniWM moves toward a more integrated formulation by unifying visual foresight and planning within a memory-augmented autoregressive architecture \cite{kim2026planning}. In parallel, EgoWM leverages pretrained internet-scale video priors for controllable egocentric prediction \cite{bagchi2026walk}, CompACT reduces planning cost through compact latent tokenization \cite{kim2026planning}, and one-step world models improve efficiency by replacing iterative diffusion with one-step generation \cite{shen2026efficient}. Despite these advances, prior work does not directly address the joint problem we consider: reducing long-horizon perceptual drift and improving geometric drift in a monocular and unified setting. Our method instead reformulates future prediction as anchor-guided rollout with bidirectional geometric grounding, which jointly mitigates recursive rollout error and strengthens geometry-aware conditioning without explicit 3D supervision.
\section{Method}

\begin{figure}[t!]
\centering
  \includegraphics[width=1.\linewidth]{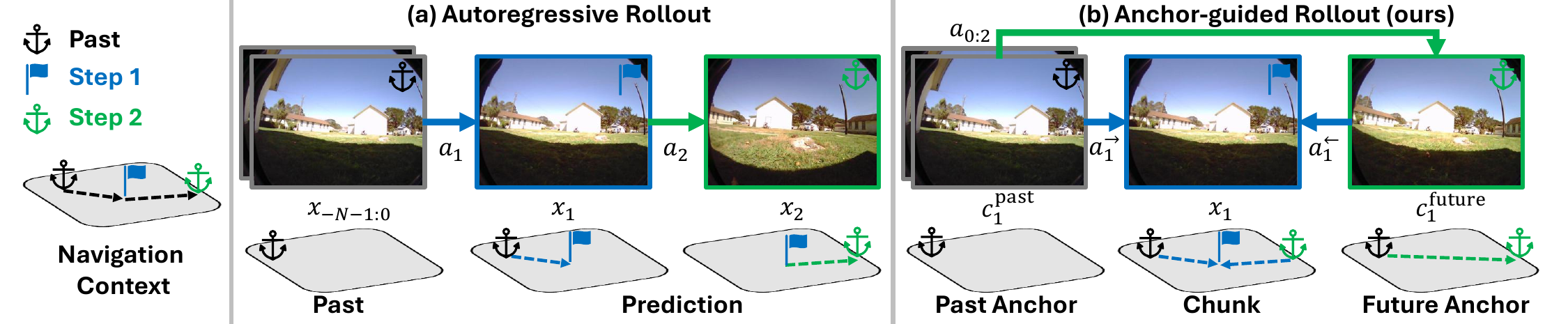}
\caption{Comparison of inference strategies.
(a) Conventional autoregressive rollout predicts future frames step by step, recursively feeding previous predictions into later inputs. (b) Our anchor-based inference first predicts sparse future anchors and then fills in the intermediate frames within each chunk under bidirectional conditions.}
   \label{fig:inference}
   \centering
\end{figure}

Conventional world models for navigation typically rely on one-sided autoregressive rollout \cite{bar2025navigation}, where previously generated outputs are recursively used as conditions for subsequent steps, as illustrated in \cref{fig:inference} (a). Formally, a world model \(p_\theta\) observes $N$ past frames $x_{s-N:s-1}$ and a possible future action $a_s$ at step $s-1$ to predict the next frame $x_s$: 
\begin{equation}
\hat{x}_{s} \sim p_\theta\!\left(x_{s}\mid \tilde{x}_{s-N:s-1}, a_s\right),
\qquad s=0,\dots,S,
\label{eq:ar}
\end{equation}
where \(S\) is the rollout horizon. In \cref{eq:ar}, \(\tilde{x}_s=x_s\) for \(s<0\), and \(\tilde{x}_s=\hat{x}_s\) for \(s\ge 0\). As rollout proceeds, predicted frames gradually replace the observed history in the conditioning window. This recursive dependency makes long-horizon prediction prone to prediction drift and geometric drift under ego-motion with accumulated noisy predictive observation $\hat{x}_s$.

\subsection{Overview}
\label{subsec:overview}

The spirit of our method is anchor-based guidance to address perceptual and geometric drifts jointly. We introduce sparse temporal anchors with two complementary roles. First, they serve as stable long-horizon targets that partition the rollout into shorter chunks, reducing recursive error accumulation (sec.~\ref{subsubsec:inference}). Second, they act as spatial reference points for bidirectional geometric masking (sec.~\ref{subsec:mask}).  This unified design is realized by AC-DiT, which generates the intermediate frames by jointly conditioning on past and future anchors, together with the geometry-aware correspondences derived from them (sec. \ref{subsubsec:model}).

\subsection{Anchor-Guided Rollout}
\label{subsubsec:inference}

To reduce the dependence on drifted intermediate predictions, we redesign the rollout order. Instead of predicting every future frame step by step, we first generate sparse future key frames, referred to as \emph{anchors}, and then generate the intermediate frames within the chunk between two anchors.

\textbf{Anchor generation.} Let
$
\mathcal{K}=\{k_1,\dots,k_M\}, 0 \le k_1 < \dots < k_M \le S,
$
denote the set of anchor steps, where adjacent anchors are separated by a fixed interval. Each anchor is predicted by directly jumping from the observation history to a farther future step using the corresponding action subsequence:
\begin{equation}
\hat{x}_{k_m}
\sim
p_\theta
\!\left(x_{k_m}\mid x_{-N+1:-1}, a_{0:k_m}\right),
\qquad m=1,\dots,M.
\end{equation}
These anchors provide coarse future anchors that capture the large-scale scene evolution under the action sequence while avoiding a long recursive rollout that repeatedly conditions on noisy intermediate predictions.

\textbf{Chunk generation.}
We then decompose the full prediction horizon into chunks between consecutive anchors. For the first chunk, the past anchor consists entirely of observed frames, while for later chunks, it is progressively updated with previously predicted anchors. Let \(k_0=0\), for each chunk \((k_{m-1},k_m)\), we generate the intermediate frames under both past and future anchors. Specifically, we define the past and future conditions as
$
c_m^{\mathrm{past}} = \hat{x}_{k_{m-1}-N-1:k_{m-1}}, 
c_m^{\mathrm{future}} = \hat{x}_{k_m}.
$
The intermediate frames inside the chunk are then generated as:
\begin{equation}
\hat{x}^{\mathrm{chunk}}_{m}
\sim
p_\theta
\!\left(
x^{\mathrm{chunk}}_{m}
\mid
c_m^{\mathrm{past}},
c_m^{\mathrm{future}},
a^{\rightarrow}_{m},
a^{\leftarrow}_{m}
\right),
\qquad m=1,\dots,M,
\end{equation}
where \(\hat{x}^{\mathrm{chunk}}_{m}\) contains the internal frames \(\hat{x}_s\) satisfying \(k_{m-1}<s<k_m\). Here, \(a^{\rightarrow}_{m}=a_{k_{m-1}:k_m}\) denotes the forward action sequence from the start of the chunk to the end, and \(a^{\leftarrow}_{m}=a_{k_m:k_{m-1}}\) denotes the corresponding inverse action sequence defined from the future anchor backward. For a target frame \(x_s\) within chunk \(m\), the forward action \(a^{\rightarrow}_{s}=a_{k_{m-1}:s}\) encodes motion from the start of the chunk to the target, while the inverse action \(a^{\leftarrow}_{s}=a_{k_m:s}\) encodes motion from the end of the chunk back to the target.

\begin{figure}[t!]
\centering
  \includegraphics[width=1.0\linewidth]{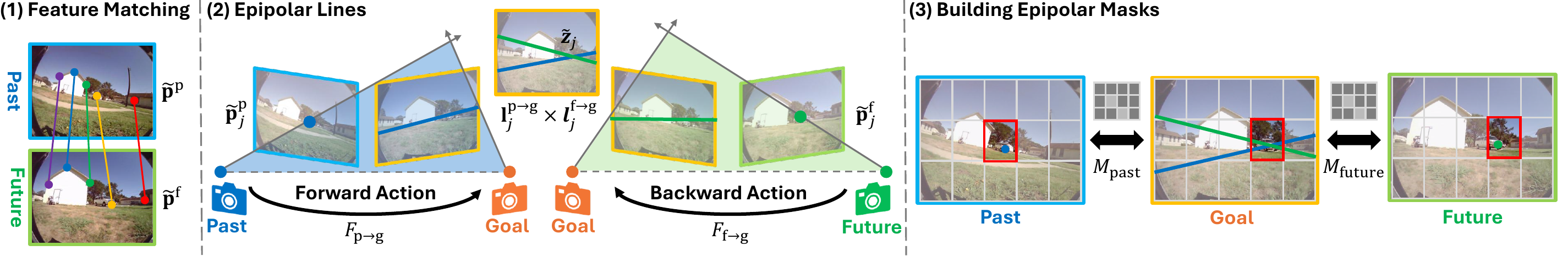}
  \caption{Bidirectional epipolar masking. (1) Match features of past and future anchors. (2)
Matched points from the past context and future anchor induce two epipolar lines in the goal frame and form an intersection. (3) This intersection is converted into attention masks that guide cross-attention toward geometry-consistent regions.}
   \label{fig:epipolar_mask}
   \centering
\end{figure}

\subsection{Bidirectional Epipolar Masking}
\label{subsec:mask}

Our masking strategy is motivated by the need to inject geometric consistency without explicit 3D priors. While the Diffusion Transformer (DiT) \cite{peebles2023scalable} is effective at generating high-quality visual content, it does not explicitly reason about scene geometry. The anchors provide an additional view-based constraint that defines regions of interest for attention, allowing the model to focus on spatially valid correspondences between the past and future views.

To implement this, we first extract sparse feature matching \cite{sun2021loftr} and estimate the fundamental matrices \cite{hartley2003multiple} \(F_{\mathrm{p}\rightarrow \mathrm{g}}\) and \(F_{\mathrm{f}\rightarrow \mathrm{g}}\) (\cref{fig:epipolar_mask}(1)). Given matched points in the past and future frames, we project them to epipolar lines in the goal frame:
\begin{equation}
\mathbf{l}^{\mathrm{p}\rightarrow \mathrm{g}}_{j} = F_{\mathrm{p}\rightarrow \mathrm{g}}\tilde{\mathbf{p}}^{\mathrm{p}}_{j},
\qquad
\mathbf{l}^{\mathrm{f}\rightarrow \mathrm{g}}_{j} = F_{\mathrm{f}\rightarrow \mathrm{g}}\tilde{\mathbf{p}}^{\mathrm{f}}_{j},
\qquad
\tilde{\mathbf{z}}_{j}
=
\mathbf{l}^{\mathrm{p}\rightarrow \mathrm{g}}_{j}\times \mathbf{l}^{\mathrm{f}\rightarrow \mathrm{g}}_{j},
\end{equation}
where \(\mathbf{l}^{\mathrm{p}\rightarrow \mathrm{g}}_{j}\) and \(\mathbf{l}^{\mathrm{f}\rightarrow \mathrm{g}}_{j}\) are the epipolar lines induced by the \(j\)-th matched points in the observation and future, respectively, and \(\tilde{\mathbf{z}}_{j}\) is their intersection in the goal  (\cref{fig:epipolar_mask} (2)). We then discretize these correspondences onto the transformer token grid and construct binary masks \(M_{\mathrm{past}}, M_{\mathrm{fut}} \in \{0,1\}^{L\times L}\), which allow attention only between geometry-consistent source and target (\cref{fig:epipolar_mask} (3)). If no valid correspondence is available for a source token, its attention row remains unmasked.

\begin{figure}[t!]
\centering
  \includegraphics[width=1.\linewidth]{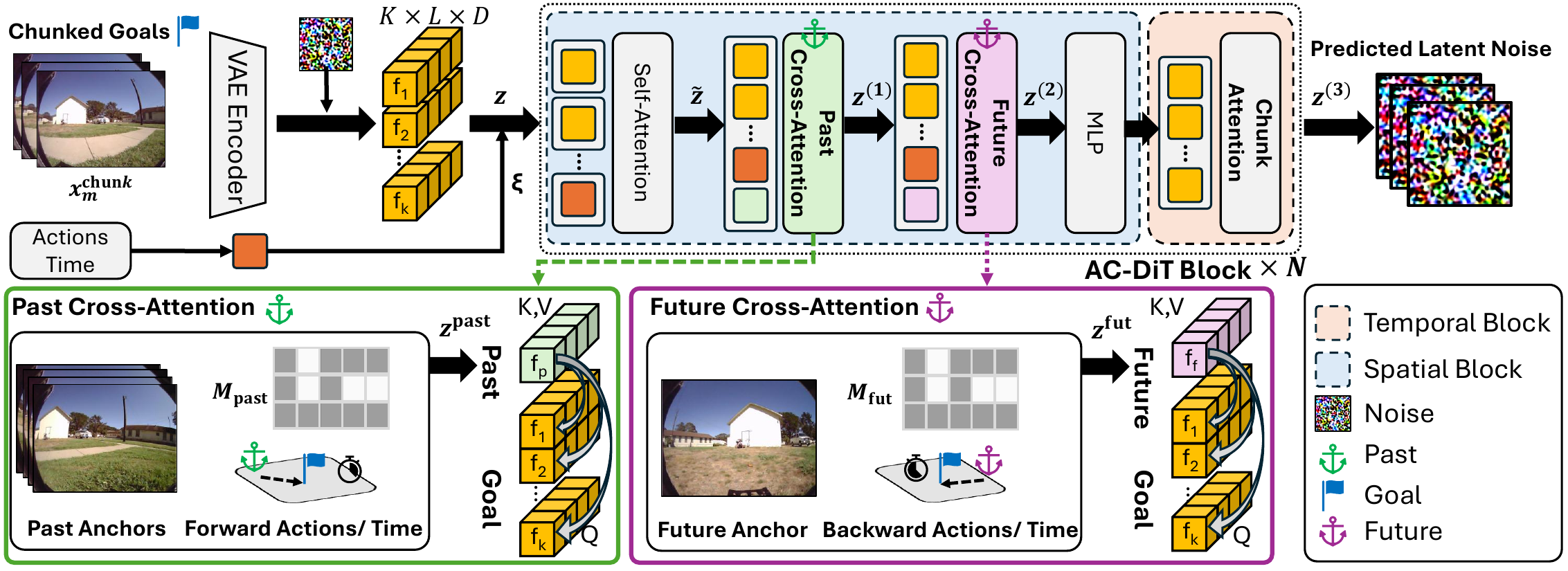}
  \caption{\textbf{Overview of AC-DiT.}
Given past and future anchors, AC-DiT generates the intermediate frames within each chunk. }
   \label{fig:model}
   \centering
\end{figure}

\subsection{Anchor-conditioned Diffusion Transformer}
\label{subsubsec:model}
AC-DiT is the generative backbone that integrates our temporal anchors and geometric constraints into a single framework. By conditioning on pairs of bidirectional anchors, the model generates the constituent frames of each chunk. As illustrated in \cref{fig:model}, the architecture extends the DiT into a multi-conditional framework that treats future prediction as a task of constrained interpolation.

\textbf{Encode bidirectional anchors.}
Scalar conditions, such as actions and time, are encoded using sinusoidal features, followed by a two-layer MLP. For the forward direction, the action \(a \in \mathbb{R}^{3}\), diffusion timestep \(t\), and relative time offset \(k\) are embedded as \(\psi_a, \psi_t, \psi_k \in \mathbb{R}^{d}\). For the backward direction, we encode the inverse action \(a^{-1}\) and future relative time \(k_f\) as \(\psi_{a^{-1}}, \psi_{k_f} \in \mathbb{R}^{d}\). The resulting scalar conditioning is:
\begin{equation}
\xi_{\mathrm{base}} = \psi_t + \psi_k + \psi_a,\qquad
\xi_{\mathrm{fut}} = \psi_{a^{-1}} + \psi_{k_f},\qquad
\xi = \xi_{\mathrm{base}} + \gamma_{\mathrm{cond}}\,\xi_{\mathrm{fut}},
\end{equation}
where \(\gamma_{\mathrm{cond}}=0\) at initialization. This preserves pretrained behavior at the beginning of finetuning while allowing the model to progressively incorporate future-side information. The term $\xi$ represents the embeddings derived from the diffusion timestep and bidirectional anchors. 

\textbf{Chunk generation with geometric guidance.}
To generate the frames within a chunk, AC-DiT processes target latent tokens $z$ by conditioning on the past anchors $z^{\mathrm{past}}$ and the future anchor $z^{\mathrm{fut}}$. Formally, for target tokens at a given diffusion step, the block computation is defined as follows. First, we use self-attention (SA) to interact with the embeddings $\xi$ denoted as $\tilde{z}=\mathrm{SA}(z;\xi)$. Next, we interact with past and future anchors through past (PA) and future cross-attention (FA):
\begin{equation}
\begin{aligned}
z^{(1)}=\tilde{z}+\gamma_{\mathrm{past}}\,\mathrm{PA}(\tilde{z},z^{\mathrm{past}};M_{\mathrm{past}},\xi), \quad
z^{(2)}&=z^{(1)}+\gamma_{\mathrm{fut}}\,\mathrm{FA}(z^{(1)},z^{\mathrm{fut}};M_{\mathrm{fut}},\xi).
\end{aligned}
\end{equation}
To interact the frames within a chunk, we use chunk attention (CA) along the temporal axis, denoted as  $z^{(3)}=z^{(2)}+\gamma_{\tau}\,\mathrm{TA}(z^{(2)};K)$, where $K$ is the chunk length. The bidirectional geometric guidance is enforced within the cross-attention layers $\mathrm{PA}$ and $\mathrm{FA}$. Specifically, the masks $M_{\mathrm{past}}$ and $M_{\mathrm{fut}}$ are applied to the attention score matrices to restrict the model’s focus to geometrically consistent regions. To ensure a stable training start, we employ zero-initialized gates $\gamma_{\mathrm{p}}$, $\gamma_{\mathrm{f}}$, and $\gamma_{\tau}$, which allow the model to gradually incorporate conditional and temporal information during the optimization.

\subsection{Training}
We train AC-DiT using the standard denoising diffusion loss \cite{ho2020denoising}. For a target chunk $x$, we minimize the mean-squared error between the added Gaussian noise $\epsilon$ and the noise predicted by our model $\epsilon_{\theta}$:
\begin{equation}
\mathcal{L} = \mathbb{E}_{x, \epsilon, t} \left[ \left| \epsilon - \epsilon_{\theta}(x^{(t)}, t, c) \right|_2^2 \right],
\end{equation}
where $x^{(t)}$ is the noisy target at step $t$ and $c$ represents the conditioning context, including the past observation, future anchor, and action sequences. This objective encourages the model to recover clean, geometry-consistent predictions from the bidirectional anchor constraints.
\section{Experiments}

\subsection{Experimental Setups}

\textbf{Datasets.}
We follow the benchmark setup of \cite{bar2025navigation} and evaluate on four navigation datasets: RECON \cite{shah2021rapid}, HuRoN \cite{hirose2023sacson} \footnote{The higher-resolution version used in the original work is not publicly available. We therefore use the publicly available lower-resolution version ($160 \times 120$).}, SCAND \cite{karnan2022socially}, and TartanDrive \cite{triest2022tartandrive}.

\textbf{Generation metrics.}
We evaluate generation quality from two complementary perspectives aligned with our main goals: \emph{perceptual drift} and \emph{geometric drift}. To quantify perceptual drift, we measure visual quality across increasing prediction horizons using FID \cite{heusel2017gans}, DreamSim \cite{fu2023dreamsim}, and LPIPS \cite{zhang2018unreasonable}, following \cite{bar2025navigation}. These metrics assess how prediction quality degrades as rollout extends farther in time. To quantify geometric, we evaluate whether generated frames remain aligned with the conditioned actions and the underlying scene geometry. Specifically, we use epipolar distance (ED) \cite{hartley2003multiple} and Sampson error (SE) \cite{sampson1982fitting,kupyn2025epipolar} to measure geometric consistency with the induced viewpoint change, and MEt3R \cite{asim2025met3r} to assess multi-view consistency after geometry-aware alignment.

\textbf{Planning metrics.}
To evaluate downstream planning benefit, we follow the planning setup of \cite{bar2025navigation}. In this setting, the planner itself is kept unchanged, so the comparison isolates the effect of improved future prediction quality. We report Absolute Trajectory Error (ATE), Final Displacement Error (FDE), and Relative Pose Error (RPE).

\textbf{Implementation details.}
We use the pretrained CDiT/XL model from \cite{bar2025navigation} as the backbone of AC-DiT. The anchor-conditioned, bidirectional, and temporal modules are initialized from scratch. Following \cite{bar2025navigation}, we adopt the VAE tokenizer from \cite{blattmann2023stable}. We fine-tune the model using the AdamW optimizer \cite{loshchilov2017decoupled} with a batch size of 4 and a learning rate of \(8\times10^{-6}\). All experiments are conducted on NVIDIA A100 GPUs. Detailed experimental setups are provided in the appendix.
\subsection{Generation}
\subsubsection{Perceptual Drift}

\begin{table*}[t]
\centering
\small
\setlength{\tabcolsep}{4pt}
\caption{Perceptual drift across horizons. Gain (\%) denotes the relative improvement over NWM.}
\label{tab:visual_full_gain}
\resizebox{\textwidth}{!}{
\begin{tabular}{ll|cccc|cccc|cccc|cccc}
\toprule
\multirow{2}{*}{} & \multirow{2}{*}{Metric}
& \multicolumn{4}{c|}{RECON}
& \multicolumn{4}{c|}{SCAND}
& \multicolumn{4}{c|}{HuRoN*}
& \multicolumn{4}{c}{TartanDrive} \\
&
& EgoWM & NWM & Ours & Gain
& EgoWM & NWM & Ours & Gain
& EgoWM & NWM & Ours & Gain
& EgoWM & NWM & Ours & Gain \\
\midrule
\multirow{3}{*}{1s}
& LPIPS $\downarrow$
& 0.504 & 0.330 & \textbf{0.281} & 14.8
& 0.497 & 0.353 & \textbf{0.297} & 15.9
& 0.653 & 0.445 & \textbf{0.334} & 24.9
& 0.481 & 0.381 & \textbf{0.303} & 20.5 \\
& DreamSim $\downarrow$
& 0.239 & 0.125 & \textbf{0.103} & 17.6
& 0.322 & 0.211 & \textbf{0.183} & 13.3
& 0.453 & 0.236 & \textbf{0.178} & 24.6
& 0.264 & 0.177 & \textbf{0.137} & 22.6 \\
& FID $\downarrow$
& 97.095 & 52.367 & \textbf{47.443} & 9.4
& 119.958 & 69.264 & \textbf{65.007} & 6.2
& 152.512 & 97.726 & \textbf{78.541} & 19.6
& 89.769 & 50.642 & \textbf{42.894} & 15.3 \\
\midrule
\multirow{3}{*}{2s}
& LPIPS $\downarrow$
& 0.591 & 0.395 & \textbf{0.325} & 17.7
& 0.524 & 0.413 & \textbf{0.345} & 16.5
& 0.676 & 0.508 & \textbf{0.389} & 23.4
& 0.527 & 0.446 & \textbf{0.373} & 16.4 \\
& DreamSim $\downarrow$
& 0.366 & 0.159 & \textbf{0.123} & 22.6
& 0.369 & 0.256 & \textbf{0.213} & 16.8
& 0.498 & 0.291 & \textbf{0.203} & 30.2
& 0.318 & 0.241 & \textbf{0.174} & 27.8 \\
& FID $\downarrow$
& 119.380 & 61.500 & \textbf{54.929} & 10.7
& 119.526 & 73.670 & \textbf{68.514} & 7.0
& 152.669 & 101.240 & \textbf{81.341} & 19.7
& 85.874 & 57.526 & \textbf{47.899} & 16.7 \\
\midrule
\multirow{3}{*}{4s}
& LPIPS $\downarrow$
& 0.651 & 0.448 & \textbf{0.385} & 14.1
& 0.601 & 0.473 & \textbf{0.401} & 15.2
& 0.733 & 0.559 & \textbf{0.439} & 21.5
& 0.642 & 0.519 & \textbf{0.439} & 15.4 \\
& DreamSim $\downarrow$
& 0.497 & 0.212 & \textbf{0.154} & 27.4
& 0.519 & 0.316 & \textbf{0.257} & 18.7
& 0.607 & 0.359 & \textbf{0.260} & 27.6
& 0.463 & 0.331 & \textbf{0.234} & 29.3 \\
& FID $\downarrow$
& 165.934 & 70.929 & \textbf{58.350} & 17.7
& 187.576 & 85.046 & \textbf{76.430} & 10.1
& 195.754 & 114.395 & \textbf{87.457} & 23.5
& 158.649 & 68.391 & \textbf{54.247} & 20.7 \\
\midrule
\multirow{3}{*}{8s}
& LPIPS $\downarrow$
& 0.727 & 0.504 & \textbf{0.444} & 11.9
& 0.680 & 0.547 & \textbf{0.464} & 15.2
& 0.774 & 0.603 & \textbf{0.506} & 16.1
& 0.748 & 0.572 & \textbf{0.504} & 11.9 \\
& DreamSim $\downarrow$
& 0.686 & 0.280 & \textbf{0.200} & 28.6
& 0.681 & 0.422 & \textbf{0.321} & 23.9
& 0.689 & 0.425 & \textbf{0.320} & 24.7
& 0.659 & 0.428 & \textbf{0.305} & 28.7 \\
& FID $\downarrow$
& 265.686 & 90.968 & \textbf{67.298} & 26.0
& 272.919 & 109.985 & \textbf{88.976} & 19.1
& 263.516 & 131.694 & \textbf{104.536} & 20.6
& 258.002 & 88.272 & \textbf{62.614} & 29.1 \\
\midrule
\multirow{3}{*}{16s}
& LPIPS $\downarrow$
& 0.765 & 0.557 & \textbf{0.508} & 8.8
& 0.719 & 0.595 & \textbf{0.510} & 14.3
& 0.787 & 0.627 & \textbf{0.544} & 13.2
& 0.796 & 0.624 & \textbf{0.553} & 11.4 \\
& DreamSim $\downarrow$
& 0.775 & 0.360 & \textbf{0.263} & 26.9
& 0.761 & 0.486 & \textbf{0.374} & 23.0
& 0.729 & 0.492 & \textbf{0.366} & 25.6
& 0.772 & 0.495 & \textbf{0.393} & 20.6 \\
& FID $\downarrow$
& 365.810 & 112.679 & \textbf{76.898} & 31.8
& 330.710 & 131.263 & \textbf{100.424} & 23.5
& 313.405 & 141.702 & \textbf{112.330} & 20.7
& 325.248 & 107.086 & \textbf{78.562} & 26.6 \\
\bottomrule
\end{tabular}
}
\end{table*}

\begin{figure}[t!]
\centering
  \includegraphics[width=1.0\linewidth]{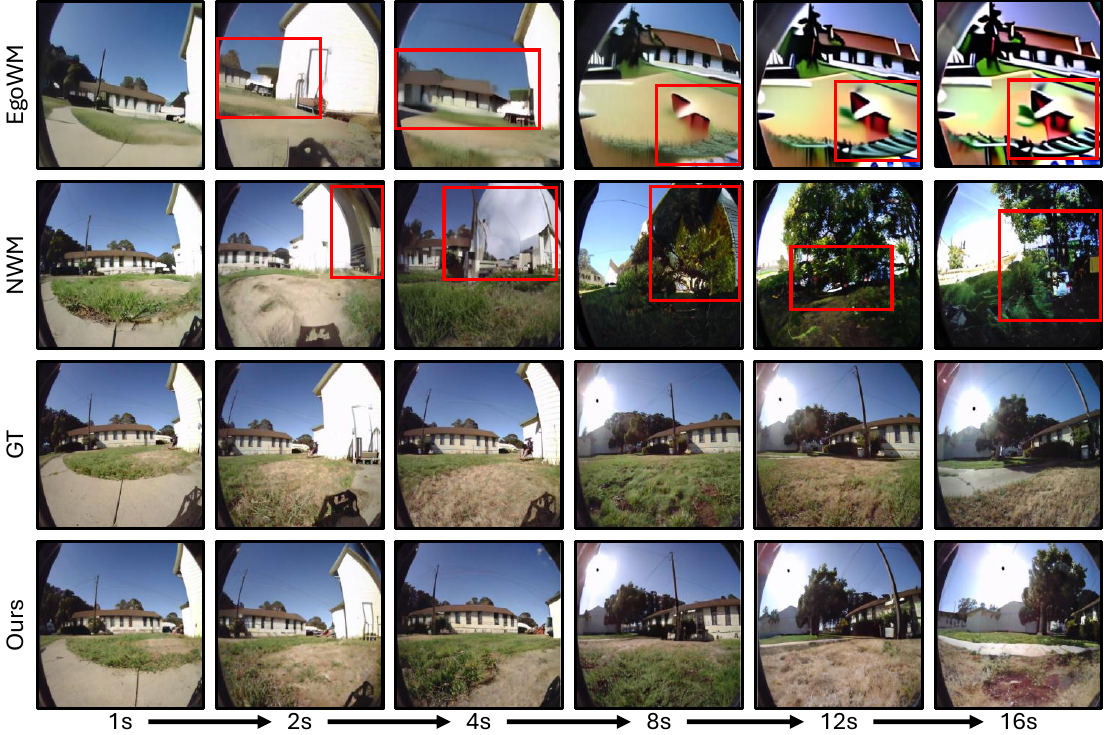}
  \caption{Qualitative results of perceptual drift over time. Red frames highlight regions where the baselines become visibly corrupted or difficult to interpret.}
   \label{fig:quality}
   \centering
\end{figure}

We evaluate perceptual drift on four datasets across horizons from \(1\)s to \(16\)s in \cref{tab:visual_full_gain}, using NWM \cite{bar2025navigation} and EgoWM \cite{bagchi2026walk} as baselines under the same \(224\times224\) image resolution. Our method consistently outperforms both baselines across all datasets, horizons, and visual-quality metrics, showing that it degrades much more gracefully as rollout extends.

The advantage becomes increasingly clear at longer horizons, where autoregressive prediction is more vulnerable to recursive error accumulation. NWM progressively deteriorates as prediction errors are fed back into later steps, while EgoWM becomes blurred and abstract after only a few seconds. In contrast, our anchor-based inference and anchor-conditioned generation reduce recursive corruption and maintain more stable long-horizon prediction. This trend is also visible in \cref{fig:quality}: NWM deteriorates noticeably beyond \(4\)s, EgoWM degrades even earlier, while our method preserves clearer structure and more interpretable content over time.

\subsubsection{Geometric Drift}

\begin{table*}[t]
\centering
\small
\setlength{\tabcolsep}{5pt}
\caption{Geometric drift comparison across datasets and prediction horizons.}
\label{tab:geometry}
\resizebox{1.0\textwidth}{!}{
\begin{tabular}{llccc|ccc|ccc|ccc}
\toprule
\multirow{2}{*}{} & \multirow{2}{*}{}
& \multicolumn{3}{c|}{RECON}
& \multicolumn{3}{c|}{SCAND}
& \multicolumn{3}{c|}{HuRoN}
& \multicolumn{3}{c}{TartanDrive} \\
& & EgoWM & NWM & Ours & EgoWM & NWM & Ours & EgoWM & NWM & Ours & EgoWM & NWM & Ours \\
\midrule
\multirow{2}{*}{1s}
& ED $\downarrow$ & 15.35 & 9.40 & \textbf{8.69} & 10.45 & 4.22 & \textbf{3.58} & 14.11 & 8.99 & \textbf{7.48} & 11.84 & 6.41 & \textbf{5.52} \\
& SE $\downarrow$  & 178.41 & 80.46 & \textbf{63.41} & 84.56 & 14.98 & \textbf{11.10} & 140.75 & 60.37 & \textbf{50.52} & 97.47 & 35.63 & \textbf{30.83} \\
\midrule
\multirow{2}{*}{2s}
& ED $\downarrow$ & 16.06 & 9.46 & \textbf{9.15} & 9.17 & 4.58 & \textbf{3.61} & 10.71 & 11.28 & \textbf{9.57} & 13.09 & 6.84 & \textbf{6.54} \\
& SE $\downarrow$  & 202.20 & 71.25 & \textbf{65.99} & 64.29 & 20.39 & \textbf{11.38} & 83.61 & 88.18 & \textbf{81.41} & 115.02 & 39.18 & \textbf{36.82} \\
\midrule
\multirow{2}{*}{4s}
& ED $\downarrow$ & 19.34 & 9.80 & \textbf{8.93} & 8.64 & 4.23 & \textbf{3.77} & 12.21 & 10.82 & \textbf{9.43} & 11.50 & 7.13 & \textbf{6.86} \\
& SE $\downarrow$  & 277.52 & 75.28 & \textbf{65.09} & 60.90 & 18.89 & \textbf{13.65} & 101.38 & 92.67 & \textbf{82.05} & 95.14 & 41.71 & \textbf{38.75} \\
\midrule
\multirow{2}{*}{8s}
& ED $\downarrow$ & 18.11 & 9.81 & \textbf{9.61} & 7.16  & 5.00 & \textbf{3.83} & 11.71 & 12.53 & \textbf{9.91} & 9.35 & 8.83 & \textbf{7.35} \\
& SE $\downarrow$  & 255.13 & 80.54 & \textbf{75.27} & 51.52 & 21.87 & \textbf{14.88} & 105.80 & 132.04 & \textbf{89.75} & 75.71 & 72.48 & \textbf{49.25} \\
\midrule
\multirow{2}{*}{16s}
& ED $\downarrow$ & 16.51 & 11.37 & \textbf{9.92} & 6.65 & 5.40 & \textbf{4.02} & \textbf{10.70} & 14.83 & 11.15 & 8.31 & 10.09 & \textbf{7.85} \\
& SE $\downarrow$  & 204.85 & 107.14 & \textbf{80.83} & 49.69 & 33.09 & \textbf{17.66} & \textbf{91.75} & 191.26 & 108.71 & 63.89 & 94.32 & \textbf{55.96} \\
\bottomrule
\end{tabular}
}
\end{table*}

\begin{figure}[t!]
\centering
  \includegraphics[width=1.0\linewidth]{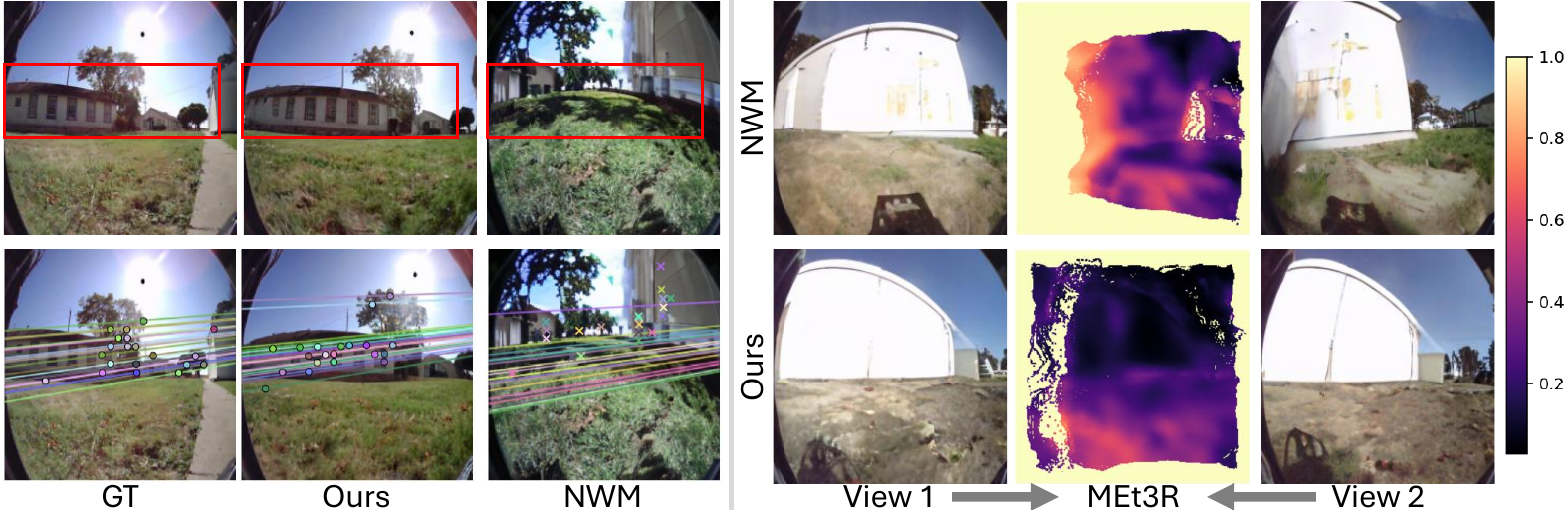}
\caption{Qualitative results of geometric drift. \textbf{Left:} Epipolar geometry visualization. Red boxes mark regions sensitive to viewpoint change under the commanded motion. In the bottom row, dots denote matches within an 8-pixel epipolar threshold, while crosses indicate larger violations. Our method better preserves scene structure and object layout, yielding more valid matches and stronger geometric alignment. \textbf{Right:} MEt3R visualization. Compared with NWM, our method preserves more coherent scene structure and viewpoint-consistent content with better MEt3R scores.}
   \label{fig:vis_geometry}
   \centering
\end{figure}

We evaluate geometric drift from two complementary perspectives: geometric alignment with the conditioned motion and multi-view consistency across viewpoint changes. In \cref{tab:geometry}, our method achieves lower epipolar distance and Sampson error than both NWM and EgoWM in nearly all settings across datasets and horizons, indicating stronger geometric consistency with the action-induced viewpoint change. The advantage becomes more pronounced at longer horizons, where autoregressive baselines accumulate more rollout error. In contrast, our method remains better aligned with the conditioned motion because future anchors provide explicit endpoint constraints and bidirectional epipolar masks further restrict attention to geometry-consistent regions. This trend is also visible in \cref{fig:vis_geometry} left, where our method preserves scene structure more faithfully and produces more geometry-consistent correspondences.

We further evaluate geometric drift under viewpoint change using MEt3R in \cref{tab:skip_met3r}. Our method consistently achieves lower MEt3R scores than NWM across all datasets and skip lengths, showing stronger multi-view coherence after geometry-aware alignment across different range of actions. As shown in \cref{fig:vis_geometry} right, our method better preserves scene structure and viewpoint consistency. Together, these results show that reducing geometric drift also improves cross-view coherence.

\begin{table*}[t]
    \centering
    \small
    
    \begin{minipage}{0.39\textwidth}
        \centering
        \caption{MEt3R comparison. Lower is better.}
        \label{tab:skip_met3r}
        \resizebox{\textwidth}{!}{
            \begin{tabular}{llccc}
                \toprule
                 & Skip & 1s &  2s & 4s \\
                \midrule
                \multirow{2}{*}{RECON}  & NWM  & 0.331 & 0.400 & 0.463 \\
                                        & Ours & \textbf{0.306} & \textbf{0.361} & \textbf{0.416} \\
                \midrule
                \multirow{2}{*}{SCAND}  & NWM  & 0.348 & 0.433 & 0.510 \\
                                        & Ours & \textbf{0.342} & \textbf{0.388} & \textbf{0.439} \\
                \midrule
                \multirow{2}{*}{HuRoN}  & NWM  & 0.466 & 0.558 & 0.628 \\
                                        & Ours & \textbf{0.429} & \textbf{0.489} & \textbf{0.540} \\
                \midrule
                \multirow{2}{*}{Tartan} & NWM  & 0.385 & 0.464 & 0.532 \\
                                        & Ours & \textbf{0.355} & \textbf{0.411} & \textbf{0.466} \\
                \bottomrule
            \end{tabular}
        }
    \end{minipage}
    \hfill
    \begin{minipage}{0.59\textwidth}
        \centering
        \caption{Planning performance.}
        \label{tab:planning}
        \resizebox{\textwidth}{!}{
            \begin{tabular}{llcccc}
                \toprule
                 & & ATE$\downarrow$ & FDE$\downarrow$ & RPE$\downarrow$ & LPIPS$\downarrow$ \\
                \midrule
                \multirow{4}{*}{RECON}  & CEM+NWM  & 3.18 & 5.86 & 0.55 & 0.438 \\
                                        & CEM+Ours & \textbf{2.78} & \textbf{5.15} & \textbf{0.51} & \textbf{0.398} \\
                \cmidrule(lr){2-6} 
                                        & NMD+NWM   & 2.97 & 5.38 & 0.52 & 0.448 \\
                                        & NMD+Ours  & \textbf{2.70} & \textbf{4.75} & \textbf{0.49} & \textbf{0.366} \\
                \midrule
                \multirow{4}{*}{SCAND}  & CEM+NWM  & 2.49 & 4.34 & 0.37 & 0.506 \\
                                        & CEM+Ours & \textbf{2.30} & \textbf{3.99} & \textbf{0.35} & \textbf{0.483} \\
                \cmidrule(lr){2-6} 
                                        & NMD+NWM   & 2.72 & 4.82 & \textbf{0.40} & 0.512 \\
                                        & NMD+Ours  & \textbf{2.69} & \textbf{4.76} & \textbf{0.40} & \textbf{0.454} \\
                \bottomrule
            \end{tabular}
        }
    \end{minipage}
\end{table*}

\subsection{Planning}

The previous results show that our method produces future predictions with lower perceptual drift and geometric drift. We next examine whether these gains translate into downstream planning benefit. Since the planner itself is kept unchanged, any performance improvement can be attributed to better predicted futures rather than changes to the planning algorithm.

Following the real-world planning setup of \cite{bar2025navigation}, we sample \(32\) candidate trajectories over a \(4\)-second horizon and report results for both  Cross-Entropy Method (CEM) \cite{rubinstein1997optimization} and NoMaD (NMD) in \cref{tab:planning}. Our method consistently improves planning performance over NWM on both datasets, indicating that more stable and action-consistent prediction leads to more accurate downstream trajectory selection.

\begin{figure}[t!]
\centering
  \includegraphics[width=1.0\linewidth]{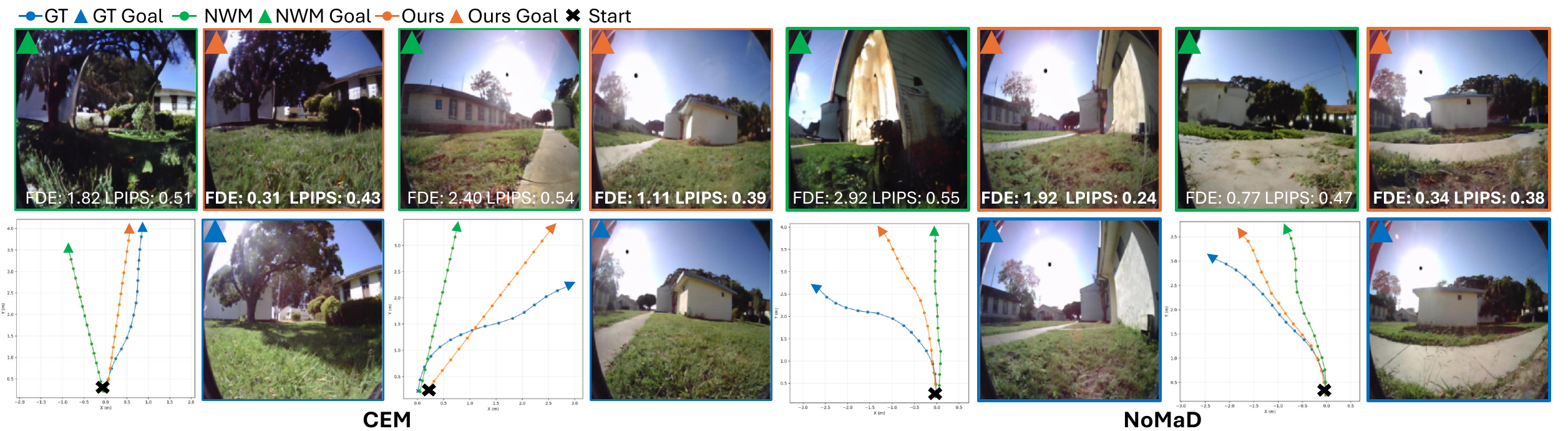}
  \caption{Qualitative planning results. Using the same planners, our method yields more coherent predictions and selects trajectories that better follow the intended motion and end closer to the goals.}
   \label{fig:planning}
   \centering
\end{figure}

This effect is also visible in \cref{fig:planning}. Because the planners operate under the same setup, the performance gap mainly comes from the quality of the predicted futures. In particular, reduced perceptual drift and geometric drift help the selected trajectory remain better aligned with the intended motion and closer to the target observation. These results show that improving future prediction quality also improves the practical usefulness of world models for downstream planning.

\subsection{Ablation Study}

We next analyze how each component contributes to the final improvements. Since our method is built on two tightly connected ideas: anchor-guided rollout and bidirectional epipolar masking. Hence, we organize the ablations according to the role of each design.

\textbf{Effect of anchor-based inference.}
We first compare conventional autoregressive rollout (AR) with the proposed anchor-based inference while keeping the same backbone model. The results in \cref{tab:ablation_inference} show that simply replacing AR with anchor-based inference already improves visual quality, confirming that the new rollout order effectively reduces recursive error accumulation and mitigates perceptual drift, but it does not directly contribute to geometric drift.

\textbf{Effect of model components.}
We next examine the contribution of each component in AC-DiT by ablating the future condition, bidirectional epipolar masking, and the chunked generation structure. The results are summarized in \cref{tab:ablation_interval}. Across variants, FID and LPIPS remain relatively similar, indicating that the main gain in long-horizon visual quality comes from the anchor-guided rollout. By contrast, the differences in Sampson error are much larger, showing that geometric drift depends strongly on the internal design of AC-DiT.

In particular, the full model achieves the best average Sampson error, indicating that all three components are needed for the strongest geometric consistency. Removing the future condition weakens the bidirectional constraint, while removing the epipolar mask removes the geometry-aware guidance. The degradation is especially clear when epipolar masking is used without future conditioning: In this case, the geometric cue degenerates from an intersection-based localization signal to a weaker line-based constraint, leading to worse epipolar consistency. These results support our main claim that bidirectional anchors and epipolar masking are complementary: The future anchor provides the second view needed for bidirectional reasoning, and the epipolar mask turns this structure into effective geometry-aware attention.

\begin{table}[t]
\centering
\small
\setlength{\tabcolsep}{4pt}
\caption{Ablation study on inference strategy. AR: autoregressive rollout. Ours: Anchored-guided rollout. }
\label{tab:ablation_inference}
\resizebox{\columnwidth}{!}{
\begin{tabular}{lccc|ccc|ccc|ccc|ccc}
\toprule
\multirow{2}{*}{}
& \multicolumn{3}{c|}{1s}
& \multicolumn{3}{c|}{2s}
& \multicolumn{3}{c|}{4s}
& \multicolumn{3}{c|}{8s}
& \multicolumn{3}{c}{16s} \\
\cmidrule(lr){2-4} \cmidrule(lr){5-7} \cmidrule(lr){8-10} \cmidrule(lr){11-13} \cmidrule(lr){14-16}
& LPIPS $\downarrow$ & FID $\downarrow$ & SE $\downarrow$
& LPIPS $\downarrow$ & FID $\downarrow$ & SE $\downarrow$
& LPIPS $\downarrow$ & FID $\downarrow$ & SE $\downarrow$
& LPIPS $\downarrow$ & FID $\downarrow$ & SE $\downarrow$
& LPIPS $\downarrow$ & FID $\downarrow$ & SE $\downarrow$ \\
\midrule
AR
& 0.333 & 53.233 & 65.76
& 0.390 & 64.975 & 66.59
& 0.451 & 78.687 & 67.75
& 0.518 & 105.839 & 79.19
& 0.577 & 140.292 & 84.74 \\
Ours
& \textbf{0.281} & \textbf{47.443} & \textbf{63.41}
& \textbf{0.325} & \textbf{54.929} & \textbf{65.99}
& \textbf{0.385} & \textbf{58.350} & \textbf{65.09}
& \textbf{0.444} & \textbf{67.298} & \textbf{75.27}
& \textbf{0.508} & \textbf{76.898} & \textbf{80.83} \\
\bottomrule
\end{tabular}
}
\end{table}

\begin{table}[t]
\centering
\small
\setlength{\tabcolsep}{6pt}
\caption{Ablation study of key components of AC-DiT. We report Sampson error under different anchor intervals, together with average LPIPS and FID. The full model consistently achieves the best geometric performance.}
\label{tab:ablation_interval}
\resizebox{1.0\columnwidth}{!}{
\begin{tabular}{ccc|cccc|c|c}
\toprule
\multirow{2}{*}{Future Anchor} & \multirow{2}{*}{Epi. Mask} & \multirow{2}{*}{Chunk}
& \multicolumn{4}{c|}{Sampson Error $\downarrow$}
& \multirow{2}{*}{Avg. LPIPS $\downarrow$}
& \multirow{2}{*}{Avg. FID $\downarrow$} \\
& & & Int.=3 & Int.=2 & Int.=1 & Avg. & & \\
\midrule
$\times$ & $\times$ & $\times$ & 67.99 & 61.24 & 53.06 & 60.76 & 0.392 & 60.040 \\
\checkmark & \checkmark & $\times$ & 65.51 & 67.50 & 51.46 & 61.49 & 0.389 & 60.312 \\
$\times$ & \checkmark & \checkmark & 76.63 & 61.76 & 54.45 & 64.31 & 0.389 & \textbf{59.447} \\
\checkmark & $\times$ & \checkmark & 66.49 & 69.24 & 52.90 & 62.88 & 0.390 & 59.456 \\
\checkmark & \checkmark & \checkmark & \textbf{63.41} & \textbf{60.89} & \textbf{44.80} & \textbf{56.37} & \textbf{0.388} & 60.983 \\
\bottomrule
\end{tabular}
}
\end{table}

\section{Conclusion}

We propose DR-NWM, a generative model that addresses two coupled failures of current approaches: perceptual drift over long autoregressive rollouts and geometric drift under ego-motion. Our method combines anchor-guided rollout, bidirectional epipolar grounding, and the AC-DiT to generate future observations that remain more stable over extended horizons while staying more faithful to the geometry induced by the agent’s actions, all without requiring explicit 3D supervision. By redesigning future prediction rollout around sparse anchors and bidirectional chunk generation, DR-NWM reduces recursive error accumulation and introduces stronger geometric constraints. Experiments on four navigation benchmarks demonstrate consistent improvements in long-horizon visual prediction, geometric consistency, multi-view coherence, and downstream planning performance under the same planners. Taken together, these results show that reducing drift and enforcing action-consistent geometric grounding are both essential for building reliable and practically useful world models for navigation. We hope this work motivates further research on world models that are not only visually plausible but also stable, geometrically grounded, and directly useful for embodied decision making.
\newpage
\section*{Acknowledgment}

This work was funded by Honda R\&D Co., Ltd, SwissAI, and Sportradar. We would like to express our gratitude to Bastien Van Delft, Mohamed Abdelfattah, and Yasamin Borhani for their insightful feedback.

\bibliographystyle{ieee_fullname} 
\bibliography{main}
\newpage
\appendix


\section{Geometric Justification of Bidirectional Epipolar Intersection}
\label{app:proof_epi}

We briefly justify why the intersection of two bidirectional epipolar lines provides a valid localization cue for the target view. Consider three views: a past view \(I^{\mathrm{past}}\), a future view \(I^{\mathrm{fut}}\), and a target (goal) view \(I^{\mathrm{goal}}\). Let \(X\) denote a 3D scene point, and let \(C_{\mathrm{past}}, C_{\mathrm{fut}}, C_{\mathrm{goal}}\) be the corresponding camera centers. Denote by \(\mathbf{x}^{\mathrm{past}}, \mathbf{x}^{\mathrm{fut}}, \mathbf{x}^{\mathrm{goal}}\) the image projections of \(X\) in the three views.

An image point does not correspond to a unique 3D point; rather, it defines a back-projected ray starting at the camera center and passing through the image plane. Thus, the point \(\mathbf{x}^{\mathrm{past}}\) defines a 3D ray \(R_{\mathrm{past}}\) through \(C_{\mathrm{past}}\), and \(\mathbf{x}^{\mathrm{fut}}\) defines a 3D ray \(R_{\mathrm{fut}}\) through \(C_{\mathrm{fut}}\). Because both image points are projections of the same scene point \(X\), the point \(X\) must lie on both rays.

Now consider the goal view. The projection of the ray \(R_{\mathrm{past}}\) onto the goal image plane defines an epipolar line \(\mathbf{l}^{\mathrm{past}\rightarrow\mathrm{goal}}\). Since \(X\in R_{\mathrm{past}}\), its projection \(\mathbf{x}^{\mathrm{goal}}\) must lie on this line. Similarly, the projection of the ray \(R_{\mathrm{fut}}\) onto the goal image plane defines another epipolar line \(\mathbf{l}^{\mathrm{fut}\rightarrow\mathrm{goal}}\), and \(\mathbf{x}^{\mathrm{goal}}\) must also lie on this line. Therefore, in the noise-free case, the target point is exactly the intersection of the two epipolar lines:
\begin{equation}
\mathbf{x}^{\mathrm{goal}}
=
\mathbf{l}^{\mathrm{past}\rightarrow\mathrm{goal}}
\cap
\mathbf{l}^{\mathrm{fut}\rightarrow\mathrm{goal}}.
\end{equation}

Using the fundamental matrix, the two epipolar lines in the goal view are given by
\begin{equation}
\mathbf{l}^{\mathrm{past}\rightarrow\mathrm{goal}}
=
F_{\mathrm{goal},\mathrm{past}}\,\tilde{\mathbf{x}}^{\mathrm{past}},
\qquad
\mathbf{l}^{\mathrm{fut}\rightarrow\mathrm{goal}}
=
F_{\mathrm{goal},\mathrm{fut}}\,\tilde{\mathbf{x}}^{\mathrm{fut}},
\end{equation}
where \(\tilde{\mathbf{x}}\) denotes homogeneous image coordinates. The target point \(\tilde{\mathbf{x}}^{\mathrm{goal}}\) satisfies both epipolar constraints,
\begin{equation}
\tilde{\mathbf{x}}^{\mathrm{goal}\top}\mathbf{l}^{\mathrm{past}\rightarrow\mathrm{goal}} = 0,
\qquad
\tilde{\mathbf{x}}^{\mathrm{goal}\top}\mathbf{l}^{\mathrm{fut}\rightarrow\mathrm{goal}} = 0.
\end{equation}
Since \(\tilde{\mathbf{x}}^{\mathrm{goal}}\) lies on both lines, it can be obtained as their cross product:
\begin{equation}
\tilde{\mathbf{x}}^{\mathrm{goal}}
=
\mathbf{l}^{\mathrm{past}\rightarrow\mathrm{goal}}
\times
\mathbf{l}^{\mathrm{fut}\rightarrow\mathrm{goal}}.
\end{equation}

In practice, estimated correspondences and fundamental matrices are noisy, so the two lines may not intersect exactly at the true target correspondence. Nevertheless, their intersection still provides a substantially tighter localization cue than the one-sided point-to-line constraint available in standard autoregressive rollout. This is the geometric basis for our bidirectional epipolar masks.

\section{Details of Bidirectional Epipolar Masking}

\subsection{Detailed Construction of Epipolar Attention Masks}
\label{app:epipolar_mask_details}

This section provides the implementation details omitted from \cref{subsec:mask}. Our goal is to use multi-view geometry to constrain cross-view attention without introducing explicit 3D reconstruction or depth supervision. The mask is constructed for each triplet consisting of a past observation, a target goal frame, and a later future anchor.

\paragraph{Triplet formation.}
During training, we use the last frame of the past context as the reference observation for geometry. For each target goal frame, we pair this observation with a future anchor sampled later in the same trajectory. In the chunked training setup used in our experiments, several consecutive goal frames share the same future anchor, which encourages temporally consistent geometric constraints across nearby targets.

\paragraph{Sparse matching and geometric estimation.}
For each triplet, we compute sparse correspondences for three image pairs: observation-goal, observation-future, and future-goal. We extract correspondences using LoFTR~\cite{sun2021loftr} and discard low-confidence matches. To improve robustness, we further reject matches that lie on unreliable regions, including sky, dynamic objects, and invalid image borders. In our implementation, this filtering combines a semantic segmentation model with lightweight color-based heuristics for sky/cloud detection and a simple border-validity test for dark fisheye boundaries.

After filtering, we estimate the fundamental matrices
\(
F_{\mathrm{p}\rightarrow \mathrm{g}}
\)
and
\(
F_{\mathrm{f}\rightarrow \mathrm{g}}
\)
with RANSAC~\cite{hartley2003multiple}. If too few filtered matches remain or RANSAC fails to produce a valid model, the corresponding pair is marked unreliable and the geometric mask is disabled for that training example.

\paragraph{Epipolar line construction.}
Let
\(
\tilde{\mathbf{p}}^{\mathrm{p}}_{j}
\)
and
\(
\tilde{\mathbf{p}}^{\mathrm{f}}_{j}
\)
denote the homogeneous coordinates of the \(j\)-th matched points in the observation and future frames, respectively. We use these points together with the estimated fundamental matrices to project each source point into the goal frame as an epipolar line:
\begin{equation}
\mathbf{l}^{\mathrm{p}\rightarrow \mathrm{g}}_{j}
=
F_{\mathrm{p}\rightarrow \mathrm{g}}
\tilde{\mathbf{p}}^{\mathrm{p}}_{j},
\qquad
\mathbf{l}^{\mathrm{f}\rightarrow \mathrm{g}}_{j}
=
F_{\mathrm{f}\rightarrow \mathrm{g}}
\tilde{\mathbf{p}}^{\mathrm{f}}_{j}.
\end{equation}
The two lines provide two independent geometric constraints on the location of the same scene element in the goal frame. We therefore compute their intersection as
\begin{equation}
\tilde{\mathbf{z}}_{j}
=
\mathbf{l}^{\mathrm{p}\rightarrow \mathrm{g}}_{j}
\times
\mathbf{l}^{\mathrm{f}\rightarrow \mathrm{g}}_{j}.
\end{equation}
If the intersection is numerically unstable or lies outside the image domain, we discard that correspondence. Otherwise, it is treated as the geometry-consistent target location in the goal frame.

\paragraph{Discretization onto the token grid.}
The geometric constraints are applied in the latent token space used by the transformer. Let the latent spatial grid contain \(L\) tokens. In our implementation, \(224\times224\) input images are encoded into a \(28\times28\) latent map and then patchified with patch size \(2\), yielding a \(14\times14\) token grid and therefore \(L=196\) spatial tokens.

Each valid observation point, future point, and goal intersection is mapped from pixel coordinates to its corresponding token index on this grid. We then build two binary masks,
\(
M_{\mathrm{past}}, M_{\mathrm{fut}} \in \{0,1\}^{L\times L},
\)
where rows correspond to source tokens and columns correspond to goal tokens. For a valid correspondence \(j\), the observation-source token is connected only to the goal token containing \(\tilde{\mathbf{z}}_{j}\) in \(M_{\mathrm{past}}\), and the future-source token is connected only to the same goal token in \(M_{\mathrm{fut}}\). If multiple valid correspondences fall into the same source token, that row may enable multiple goal tokens. If no valid correspondence exists for a source token, its row remains fully unmasked, allowing the model to revert to unconstrained attention rather than enforcing an incorrect geometric hypothesis.

\paragraph{Reliability-aware gating.}
To avoid using noisy geometry, we compute a reliability score for each pairwise matching result from the number of RANSAC inliers. For a pair \(a\!\rightarrow\!b\), we define
\begin{equation}
r_{a\rightarrow b}
=
\mathbbm{1}[F_{a\rightarrow b}\ \text{exists}]
\cdot
\mathrm{clip}
\!\left(
\frac{n_{a\rightarrow b}-n_{\min}}{n_{\mathrm{sat}}-n_{\min}},
0,1
\right),
\end{equation}
where \(n_{a\rightarrow b}\) is the number of inlier correspondences, \(n_{\min}\) is the minimum inlier count required before the score becomes non-zero, and \(n_{\mathrm{sat}}\) is the count at which the score saturates to \(1\). The final triplet reliability is the minimum of the observation-goal, observation-future, and future-goal scores:
\begin{equation}
r_{\mathrm{triplet}}
=
\min
\bigl(
r_{\mathrm{p}\rightarrow \mathrm{g}},
r_{\mathrm{p}\rightarrow \mathrm{f}},
r_{\mathrm{f}\rightarrow \mathrm{g}}
\bigr).
\end{equation}
If \(r_{\mathrm{triplet}}\) falls below a threshold, we disable the geometric mask and revert to full attention for that example.

\paragraph{Temporal smoothing in chunked training.}
When several consecutive goal frames are trained jointly as a temporal chunk, the mask can vary abruptly from frame to frame due to noisy sparse matches. To reduce this jitter, we smooth the binary masks over time with an exponential moving average. If \(M_t\) denotes the binary mask for the \(t\)-th frame in a chunk, we compute
\begin{equation}
\bar{M}_t
=
\alpha \bar{M}_{t-1}
+
(1-\alpha) M_t,
\end{equation}
followed by thresholding to recover a binary mask. This encourages neighboring goal frames in the same chunk to use consistent geometric constraints.

\paragraph{Use inside the transformer.}
The resulting masks are applied as cross-attention masks, rather than as an auxiliary loss. Specifically, they constrain which goal tokens may attend to observation and future tokens during cross-view conditioning. In this way, the model retains the expressive power of DiT while being biased toward spatially valid correspondences supported jointly by the observation and future anchor.

\subsection{Hyperparameters}
 In the chunked training setup, we use a chunk length of \(K=3\), so that each training sample contains short temporal windows over which the geometric and temporal consistency constraints are enforced. For sparse matching, we retain LoFTR \cite{sun2021loftr} correspondences with confidence above \(\tau_{\mathrm{match}}=0.8\). We then estimate the fundamental matrices with RANSAC using a reprojection threshold of \(\tau_{\mathrm{ransac}}=3\) pixels, and only attempt geometric estimation when at least 8 filtered matches are available. For the reliability-aware mask gating described in \cref{app:epipolar_mask_details}, we set the minimum inlier count to \(n_{\min}=16\), the saturation count to \(n_{\mathrm{sat}}=64\), and the reliability threshold to \(\tau_{\mathrm{rel}}=0.1\). For chunked mask stabilization, we smooth the binary masks temporally with an exponential moving average using \(\alpha=0.6\), and then binarize the smoothed masks with threshold \(\tau_{\mathrm{temp}}=0.5\). Unless otherwise stated, we use the point-intersection version of the epipolar mask.

\section{Detailed Experimental Setups}
\label{app:exp_details}

\subsection{Datasets}

We follow the benchmark protocol of \cite{bar2025navigation} and evaluate on four navigation datasets: RECON \cite{shah2021rapid}, HuRoN \cite{hirose2023sacson}, SCAND \cite{karnan2022socially}, and TartanDrive \cite{triest2022tartandrive}. Together, these datasets cover a diverse range of navigation scenarios, including real-world indoor and outdoor robot navigation as well as large-scale synthetic driving environments. This diversity is important for evaluating both long-horizon prediction quality and action-conditioned viewpoint consistency under different scene layouts, motion patterns, and visual appearances.

\textbf{RECON}
 is a real-world robotic navigation dataset designed for visual navigation and exploration. It contains egocentric image observations paired with robot motion, making it suitable for evaluating action-conditioned future prediction in realistic embodied settings. Its real-world scene diversity and viewpoint changes make it a useful benchmark for studying long-horizon rollout stability.

\textbf{HuRoN}
 is a real-world navigation dataset collected in human-centered environments. Compared with standard robot-navigation benchmarks, it contains more frequent scene changes caused by clutter, pedestrians, and dynamic surroundings, which makes future prediction and action-conditioned consistency more challenging. Following \cite{bar2025navigation}, we use the publicly available lower-resolution version, since the higher-resolution version used in the original work is not publicly available.

\textbf{SCAND}
 focuses on socially aware navigation in indoor environments. The dataset includes robot trajectories collected in the presence of people and obstacles, introducing complex viewpoint changes and interaction-rich motion patterns. These properties make SCAND particularly relevant for evaluating whether predicted futures remain aligned with commanded motion in crowded or structured environments.

\textbf{TartanDrive}
 is a large-scale driving dataset that provides egocentric observations under diverse outdoor conditions. Compared with the indoor robot-navigation datasets above, TartanDrive involves larger scene depth, faster ego-motion, and longer-range viewpoint changes. It therefore provides a complementary testbed for evaluating prediction drift and action-conditioned coherence in more challenging long-horizon settings.

\subsection{Baselines}
For NWM \cite{bar2025navigation}, we use the CDiT/XL checkpoint released by the authors as our baseline. This checkpoint is trained on the additional Ego4D dataset to provide better generalizability. For EgoWM \cite{bagchi2026walk}, we use the official pretrained Stable Video Diffusion (SVD) weights \cite{blattmann2023stable}. To ensure a fair comparison, we directly generate images at the same resolution as NWM.

\subsection{Metrics}

\paragraph{Epipolar Distance.}
We measure epipolar distance as the pixel distance between a matched point and its corresponding epipolar line induced by the estimated fundamental matrix. For a correspondence \(x \leftrightarrow x'\) and a fundamental matrix \(F\), the epipolar line in the second view is
\begin{equation}
\mathbf{l}' = F x = (a,b,c)^\top.
\end{equation}
The epipolar distance from the point \(x'=(u',v',1)^\top\) to this line is
\begin{equation}
d_{\mathrm{epi}}(x', \mathbf{l}')
=
\frac{|a u' + b v' + c|}{\sqrt{a^2+b^2}}.
\end{equation}
We compute this quantity for valid correspondences and report the average over pairs. Lower values indicate that generated observations remain more consistent with the conditioned viewpoint change. The point-to-line mapping induced by \(F\) and the standard point-to-line distance are classical epipolar-geometry constructions. 
\paragraph{Sampson Error.}
We also report the Sampson epipolar error, a first-order approximation to geometric reprojection error:
\begin{equation}
\mathrm{SE}
=
\frac{(x'^{\top} F x)^2}
{(Fx)_1^2 + (Fx)_2^2 + (F^{\top}x')_1^2 + (F^{\top}x')_2^2}.
\end{equation}
Lower Sampson error indicates better satisfaction of the epipolar constraint and stronger geometric consistency in the generated observations.

\paragraph{MEt3R.}
We use MEt3R \cite{asim2025met3r} as a metric of multi-view consistency. Given two images \(I_i\) and \(I_j\), MEt3R first applies DUSt3R \cite{wang2024dust3r} to infer a dense shared 3D reconstruction from the image pair. This reconstruction is used to warp one image into the viewpoint of the other, after which the aligned views are compared in feature space rather than pixel space. Abstractly, the metric can be written as
\begin{equation}
\mathrm{MEt3R}(I_i,I_j)=D\!\left(\phi\!\left(W_{i\rightarrow j}(I_i)\right),\,\phi(I_j)\right),
\end{equation}
where \(W_{i\rightarrow j}(\cdot)\) denotes geometry-aware warping, \(\phi(\cdot)\) denotes the feature extractor, and \(D(\cdot,\cdot)\) denotes the feature-space discrepancy used by the metric. Because it evaluates consistency after geometry-aware alignment, MEt3R is well-suited for measuring whether generated observations remain coherent under viewpoint change. Lower values indicate stronger multi-view consistency. 

\paragraph{Absolute Trajectory Error (ATE).}
ATE measures the overall deviation between the predicted trajectory and the ground-truth trajectory over the full horizon. Let \(\hat{\mathbf{p}}_t\) and \(\mathbf{p}_t\) denote the predicted and ground-truth positions at time step \(t\), respectively. We compute ATE as
\begin{equation}
\mathrm{ATE}
=
\frac{1}{T}\sum_{t=1}^{T}
\left\|
\hat{\mathbf{p}}_t - \mathbf{p}_t
\right\|_2.
\end{equation}
Lower ATE indicates that the predicted trajectory remains closer to the ground-truth path throughout the horizon.

\paragraph{Final Displacement Error (FDE).}
FDE measures the endpoint error of the predicted trajectory. It is defined as the Euclidean distance between the predicted final position and the ground-truth final position:
\begin{equation}
\mathrm{FDE}
=
\left\|
\hat{\mathbf{p}}_{T} - \mathbf{p}_{T}
\right\|_2.
\end{equation}
Lower FDE indicates better accuracy at the final destination.

\paragraph{Relative Pose Error (RPE).}
RPE measures the local consistency of the predicted trajectory by comparing the relative motion between consecutive steps. Let
\(\Delta \hat{\mathbf{p}}_t = \hat{\mathbf{p}}_t - \hat{\mathbf{p}}_{t-1}\)
and
\(\Delta \mathbf{p}_t = \mathbf{p}_t - \mathbf{p}_{t-1}\)
denote the predicted and ground-truth relative displacements, respectively. We define RPE as
\begin{equation}
\mathrm{RPE}
=
\frac{1}{T-1}\sum_{t=2}^{T}
\left\|
\Delta \hat{\mathbf{p}}_t - \Delta \mathbf{p}_t
\right\|_2.
\end{equation}
Lower RPE indicates better preservation of local motion consistency along the trajectory.

\section{Additional Ablations}
\textbf{Effect on downstream planning.}
Finally, we analyze when the improved world model is most beneficial for downstream planning. \cref{tab:samples} studies the effect of the number of sampled trajectories. Our method consistently improves ATE over NWM, with the largest gains appearing when the number of samples is limited. As the number of samples increases, the gap gradually narrows, suggesting that better future prediction is especially helpful in low-sample regimes where the planner cannot rely on exhaustive search. When the number of samples becomes very large, the planner already covers a wide range of possible trajectories, reducing the relative importance of improved ranking from the world model.
\begin{table}[t]
\centering
\small
\setlength{\tabcolsep}{8pt}
\caption{Effect of the number of samples on ATE. Lower is better.}
\label{tab:samples}
\resizebox{0.5\columnwidth}{!}{
\begin{tabular}{lcccc}
\toprule
Samples & 10 & 20 & 30 & 60 \\
\midrule
CEM + NWM  & 3.93 & 3.28 & 3.17 & 2.97 \\
CEM + Ours & \textbf{3.69} & \textbf{3.06} & \textbf{2.78} & \textbf{2.71} \\
\bottomrule
\end{tabular}
}
\end{table}

\section{Additional Qualitative Results}

\subsection{Perceptual Drift}
\begin{figure}[t]
\centering
  \includegraphics[width=0.8\linewidth]{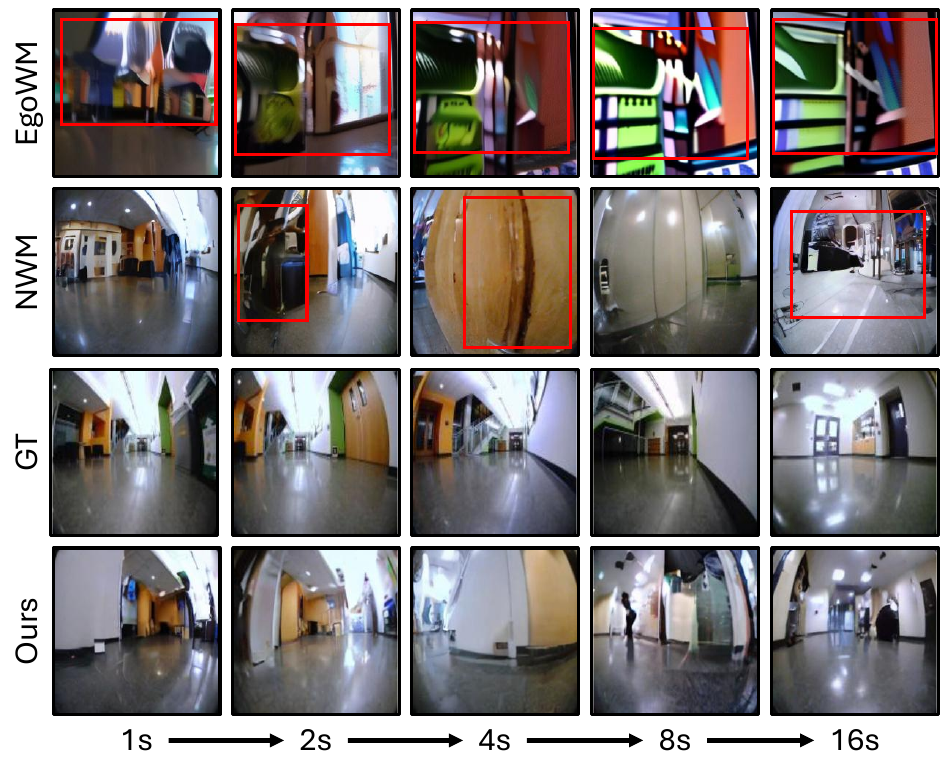}
  \caption{\textbf{Qualitative results over time on HuRoN.} As the prediction horizon increases, NWM and EgoWM exhibit stronger blur, drift, and structural degradation, while our method preserves clearer layouts and more stable object appearance over longer rollouts. Red frames highlight regions where the baselines become visibly corrupted or difficult to interpret.}
   \label{fig:quality_huron}
   \centering
\end{figure}

\begin{figure}[t]
\centering
  \includegraphics[width=0.8\linewidth]{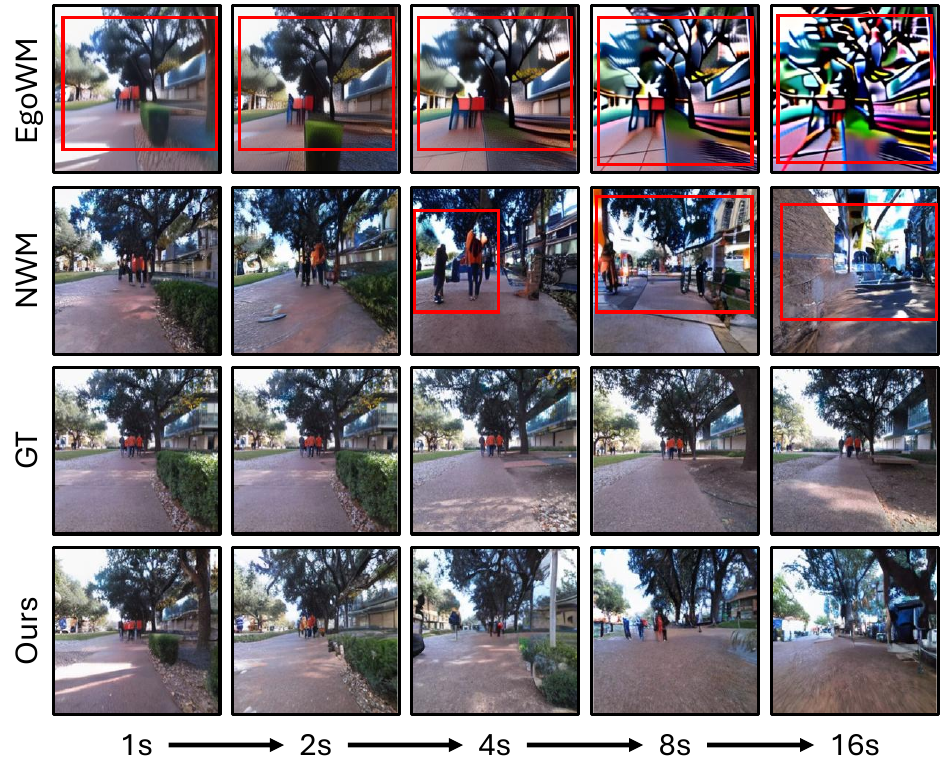}
  \caption{\textbf{Qualitative results over time on SCAND.} Compared with the baselines, our method better preserves scene structure and viewpoint consistency as the rollout extends, showing reduced perceptual drift over long horizons. Red frames highlight regions where the baselines become visibly corrupted or difficult to interpret.}
   \label{fig:quality_scand}
   \centering
\end{figure}

\begin{figure}[t]
\centering
  \includegraphics[width=0.8\linewidth]{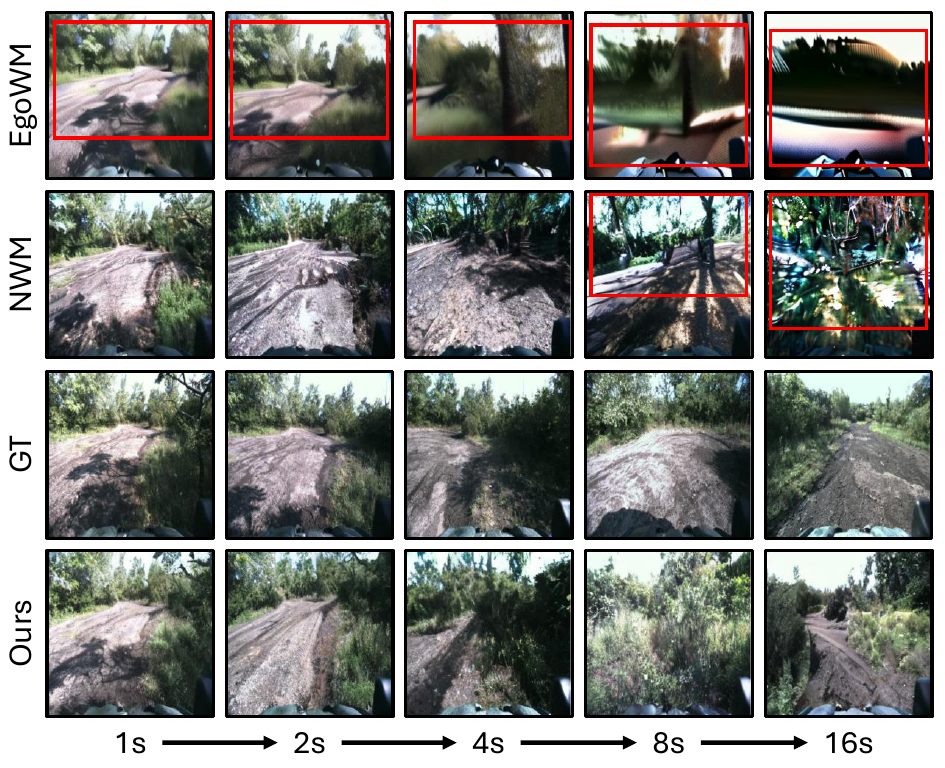}
  \caption{\textbf{Qualitative results over time on TartanDrive.} In this more challenging driving setting, our method maintains more stable long-horizon predictions, whereas the baselines deteriorate more noticeably as the horizon increases. Red frames highlight regions where the baselines become visibly corrupted or difficult to interpret.}
   \label{fig:quality_tartan}
   \centering
\end{figure}

We provide additional qualitative comparisons on HuRoN (\cref{fig:quality_huron}), SCAND (\cref{fig:quality_scand}), and TartanDrive (\cref{fig:quality_tartan}) to illustrate how prediction quality evolves as the rollout horizon increases.

\subsection{Geometry Drift}
\begin{figure}[t]
\centering
  \includegraphics[width=0.9\linewidth]{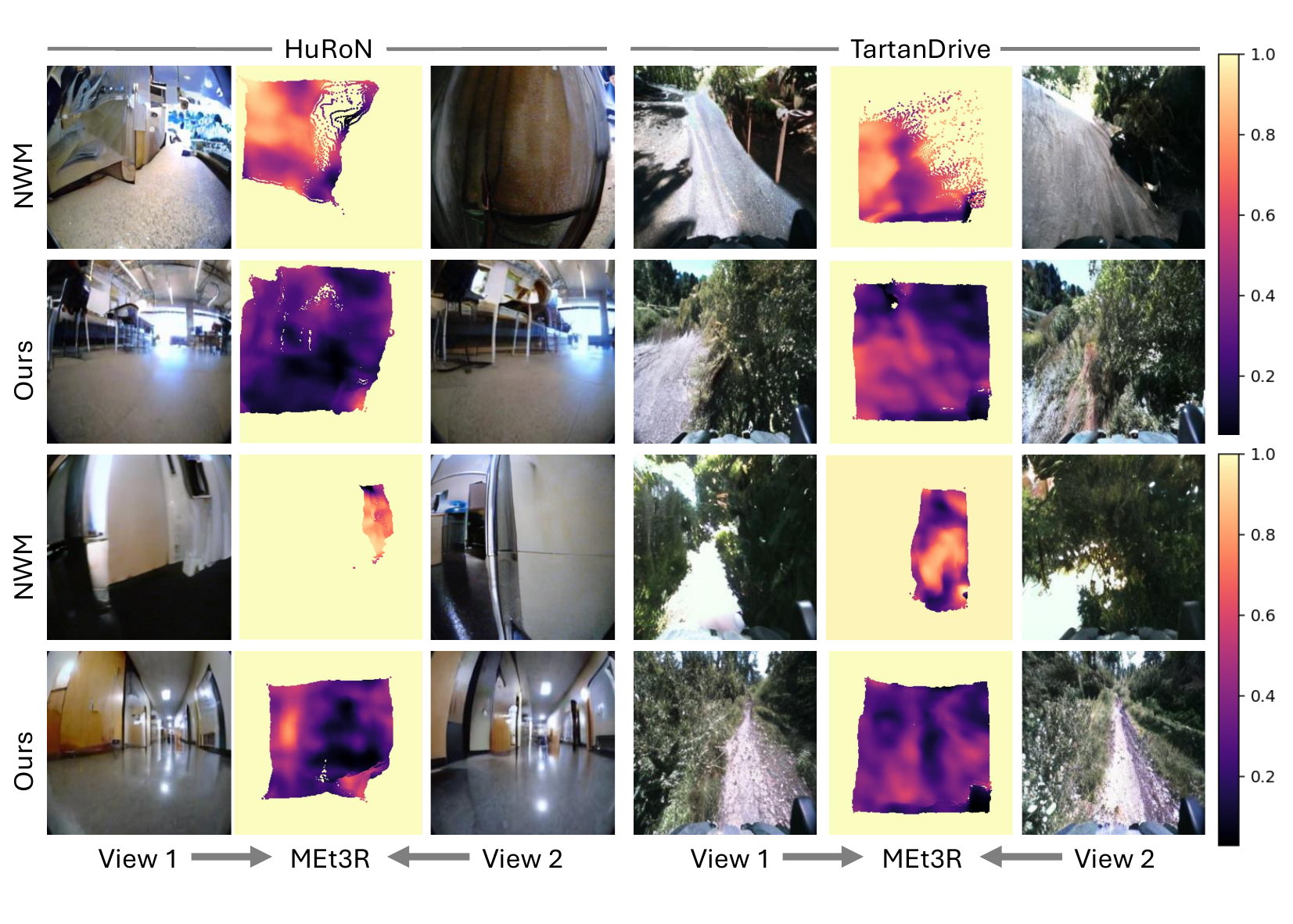}
  \caption{\textbf{Qualitative results of MEt3R on HuRoN and TartanDrive.} We visualize multi-view inconsistency measured by MEt3R after geometry-aware alignment. Darker responses indicate lower inconsistency and therefore stronger cross-view coherence. Our method preserves more coherent scene structure and viewpoint-consistent content.}
   \label{fig:quality_met3r_1}
   \centering
\end{figure}

\begin{figure}[t]
\centering
  \includegraphics[width=0.9\linewidth]{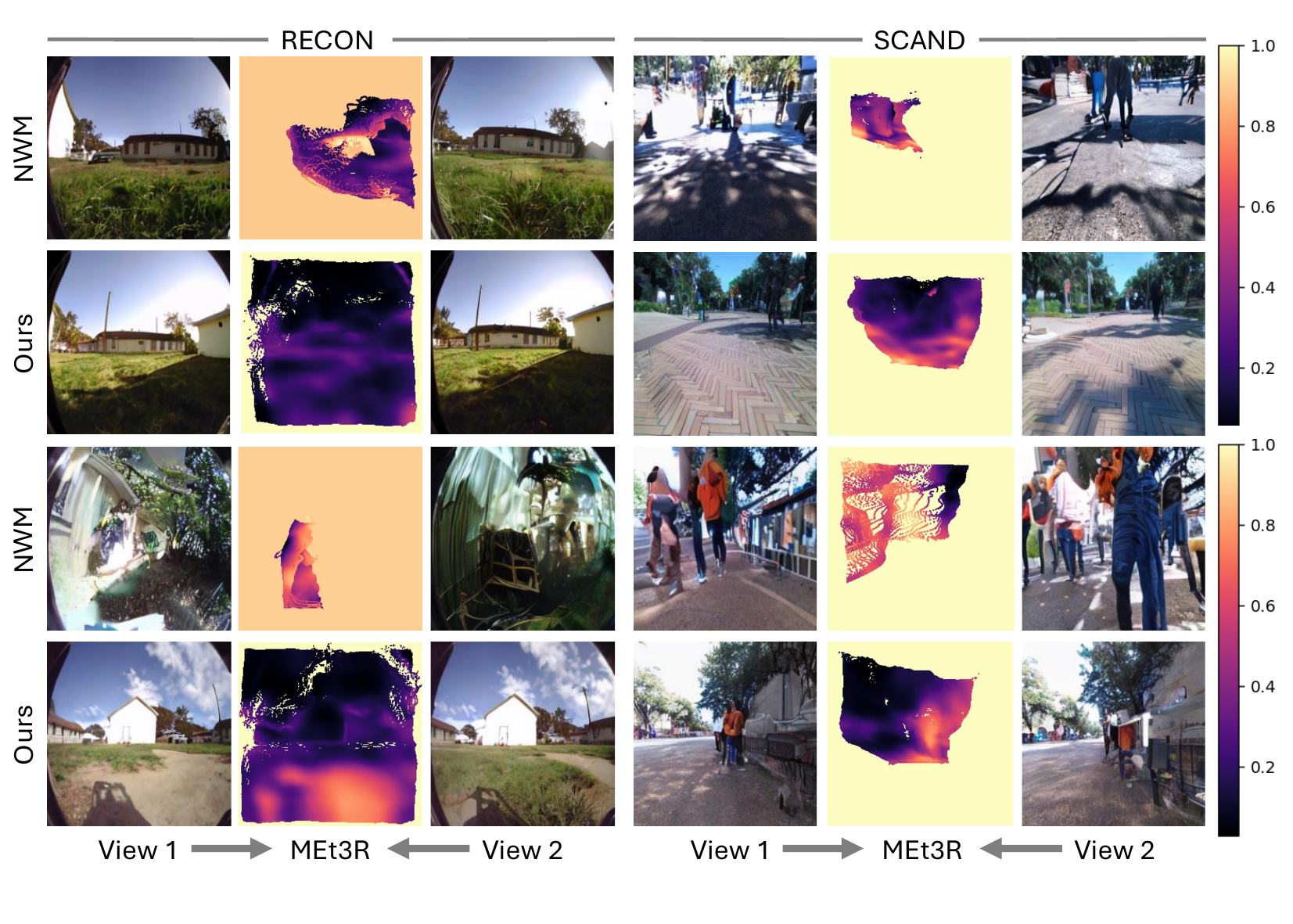}
  \caption{\textbf{Qualitative results of MEt3R on RECON and SCAND.} Across different scenes, our method produces more consistent cross-view content after alignment, while the baseline shows stronger geometric inconsistency and structural mismatch.}
   \label{fig:quality_met3r_2}
   \centering
\end{figure}

We further visualize geometry drift through additional MEt3R in \cref{fig:quality_met3r_1} and \cref{fig:quality_met3r_2}, which highlight the difference in multi-view consistency between our method and the baseline under viewpoint changes.

\end{document}